\documentclass[a4paper,twoside]{article}

\usepackage{epsfig}
\usepackage{subcaption}
\usepackage{calc}
\usepackage{amssymb}
\usepackage{amstext}
\usepackage{amsmath}
\usepackage{amsthm}
\usepackage{multicol}
\usepackage{pslatex}
\usepackage{apalike}
\usepackage[bottom]{footmisc}

\usepackage{comment}
\usepackage{todonotes}
\usepackage{SCITEPRESS}     
\usepackage[pagebackref=true,breaklinks=true,colorlinks,bookmarks=false]{hyperref}

\DeclareMathOperator*{\argmax}{arg\, max}
\newcommand{\IoU}{\mathit{IoU}}
\newcommand{\mIoU}{\mathit{mIoU}}
\newcommand{\auroc}{\mathit{AUROC}}
\newcommand{\auc}{\mathit{AUPRC}}

\newcommand{\rec}{\mathit{REC}_{80}}

\begin{document}

\title{False Negative Reduction in Semantic Segmentation \\ under Domain Shift using Depth Estimation}

\author{ }
\author{\authorname{Kira Maag\sup{1} and Matthias Rottmann\sup{2,3}}
\affiliation{\sup{1}Ruhr University Bochum, Germany \hspace{2ex} \sup{2}University of Wuppertal, Germany \hspace{2ex} \sup{3} EPFL, Switzerland}
\email{kira.maag@rub.de, rottmann@uni-wuppertal.de}
}


\abstract{State-of-the-art deep neural networks demonstrate outstanding performance in semantic segmentation. However, their performance is tied to the domain represented by the training data. Open world scenarios cause inaccurate predictions which is hazardous in safety relevant applications like automated driving. In this work, we enhance semantic segmentation predictions using monocular depth estimation to improve segmentation by reducing the occurrence of non-detected objects in presence of domain shift. To this end, we infer a depth heatmap via a modified segmentation network which generates foreground-background masks, operating in parallel to a given semantic segmentation network. Both segmentation masks are aggregated with a focus on foreground classes (here road users) to reduce false negatives. To also reduce the occurrence of false positives, we apply a pruning based on uncertainty estimates. Our approach is modular in a sense that it post-processes the output of any semantic segmentation network. In our experiments, we observe less non-detected objects of most important classes and an enhanced generalization to other domains compared to the basic semantic segmentation prediction.}

\onecolumn \maketitle \normalsize \setcounter{footnote}{0} \vfill

\section{\uppercase{Introduction}}
\begin{figure}[t]
    \center
    \includegraphics[width=0.47\textwidth]{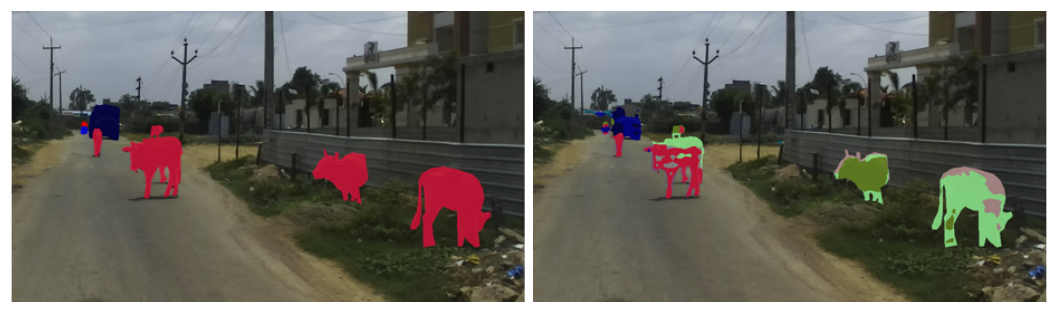}
    \caption{Example image of the India Driving dataset. \emph{Left}: Ground truth pixels of classes humans/animals colored in red and vehicles in blue. \emph{Right}: Semantic segmentation for the mentioned classes.}
    \label{fig:idd}
\end{figure}

Semantic image segmentation aims at segmenting objects in an image by assigning each pixel to a class within a predefined set of semantic classes. Thereby, semantic segmentation provides comprehensive and precise information about the given scene. This is particularly desirable in safety relevant applications like automated driving. In recent years, deep neural networks (DNNs) have demonstrated outstanding performance on this task \cite{Chen2018,Wang2021}. 
However, DNNs are usually trained on a specific dataset (source domain) and often fail to function properly on unseen data (target domain) due to a domain gap. 
In real-world applications, domain gaps may occur due to shifts in location, time and other environmental parameters. This causes domain shift on both, foreground classes – countable objects such as persons, animals, vehicles – and background classes – regions with similar texture or material like sky, road, nature, buildings \cite{Adelson20O1}.
\autoref{fig:idd} gives an example for the lack of generalization, i.e., the DNN is trained on street scenes in German cities \cite{Cordts2016} resulting in defective behavior on the unseen India road scenes \cite{Varma2019} where the animals are predicted as person, nature or fence.
This is critical since potential hazardous situations are underestimated due to the prediction of non-dynamic classes. 
On the one hand, when using semantic segmentation in open world scenarios, the appearance of objects that do not belong to any of the semantic classes the DNN has been trained on (like animals) may cause defective predictions \cite{Pinggera2016}. On the other hand, even objects of known classes can change their appearance, leading to erroneous predictions. Hence, for the deployment of DNNs in safety-critical applications, robustness under domain shifts is essential.

\emph{Unsupervised domain adaptation} is an approach overcoming this issue. The idea is to train a DNN on labeled source domain data and jointly on unlabeled target data adapting the source domain distribution to the target one \cite{Watanabe2018}. As target data is not always available for training, the recent research has also been devoted to \emph{domain generalization} resolving this limitation \cite{Lee2022}.

\begin{figure*}[t]\center
\center
\includegraphics[trim=0 0 0 3,clip,width=0.9\textwidth]{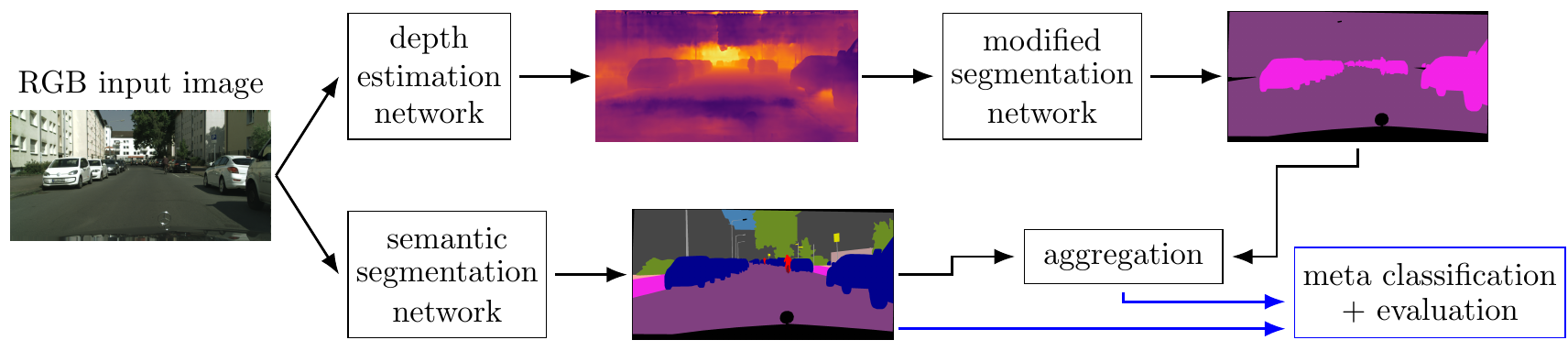}
\caption{Overview of our method. The input image is fed into a semantic segmentation network (bottom branch) and in parallel into a depth estimation network (top branch). The resulting depth heatmap is passed to our modified segmentation network which predicts a foreground-background segmentation. This prediction is aggregated with the semantic segmentation and finally, meta classification is applied to reduce false positive segments in an uncertainty-aware manner.}
    \label{fig:overview}
\end{figure*}

In this work, we introduce a domain generalization method for semantic segmentation using depth estimation focusing on the reduction of false negative foreground objects. In applications like automated driving, the foreground class is of particular interest due to its dynamical behavior. Especially in presence of domain gaps, the detection performance w.r.t.\ these object classes can decrease significantly.
An overview of our approach consisting of two branches (running in parallel) is shown in \autoref{fig:overview}.
The \emph{image segmentation branch} is a semantic segmentation inference and the \emph{depth segmentation branch} feeds the same RGB input image into a depth estimation network. The goal of depth estimation is to obtain a representation of the spatial structure of a given scene, which can help to bridge domain gaps \cite{Wang2021_0}. The resulting depth heatmap is passed to a modified segmentation network which predicts foreground-background segmentation. The architecture of this network may be based on the architecture of the semantic segmentation network, but can be chosen independently.
In the \emph{fusion} step, the semantic segmentation and the foreground-background prediction are aggregated obtaining several segments (connected components of pixels belonging to the same class) per foreground class. As a result of combining the two masks, we detect overlooked segments of the basic semantic segmentation network on the source dataset as well as under domain shift using the depth information for domain generalization. However, the increased sensitivity towards finding foreground objects may result in an overproduction of false positive segments. To overcome this, we utilize an uncertainty-aware post-processing fusion step, a so-called \emph{meta classifier} which performs \emph{false positive pruning} with a lightweight classifier \cite{Rottmann2018,Maag2019}. Moreover, to gain a further performance boost, the meta classifier, which is trained only on the source domain, can be fine-tuned on a small amount of the respective target domain (lightweight domain adaptation).

We only assume input data as well as a trained semantic segmentation and a depth estimation network. Due to the modularity of our method, we can set up our model based on these assumptions and it is applicable to any semantic segmentation network i.e., only the output is post-processed. In our tests, we employ two semantic segmentation \cite{Chen2018,Zhang2019} and two depth estimation networks \cite{Godard2019,Lee2019} applied to four datasets, i.e., Cityscapes \cite{Cordts2016} as source domain and A2D2 \cite{Geyer2020}, LostAndFound \cite{Pinggera2016} as well as India Driving \cite{Varma2019} as target domains. 
The application of these widely differing datasets is intended to demonstrate the domain generalization and error reduction capability of our approach. 
The source code is publicly available at \url{http://github.com/kmaag/FN-Reduction-using-Depth}.
Our contributions are summarized as follows:
\begin{itemize}
    \item We introduce a modified segmentation network which is fed with depth heatmaps and outputs foreground-background segmentation masks which are combined with semantic segmentation masks to detect possible overlooked segments (by the semantic segmentation network) of the most important classes. In addition, we perform meta classification to prune false positive segments in an uncertainty-aware fashion.
    \item For the first time, we demonstrate that incorporating depth information in a post-processing step improves a semantic segmentation performance (independently of the choice of semantic segmentation network). We compare the performance of our method with basic semantic segmentation performance on several datasets (with domain gap) obtaining area under precision-recall curve values of up to $97.08\%$ on source domain and $93.83\%$ under domain shift. 
\end{itemize}
The paper is structured as follows. In \autoref{sec:related_work}, we discuss the related work. Our approach is introduced in \autoref{sec:method} including the modified segmentation network, the aggregation of network predictions and meta classification. The numerical results are shown in \autoref{sec:result}.
%
%
\section{\uppercase{Related Work}}\label{sec:related_work}
In this section, we first discuss related methods improving robustness of DNNs under domain shift as well as false negative reduction approaches. Thereafter, we present works that use depth information to enhance semantic segmentation prediction.
%
%
\paragraph{Robustness under Domain Shift} 
Unsupervised domain adaptation is often used to strengthen the robustness of DNNs bridging domain gaps \cite{Watanabe2018}. The DNN is trained with source data (labeled) and target data (unlabeled and different from source dataset) to align the target domain's distributions. In \cite{Yan2019}, this problem is tackled by a generative adversarial network which translates the target domain into the source domain before predicting semantic segmentation. Monocular depth estimation is used in \cite{Cardace2022,Wang2021_0} to improve the prediction performance under domain shift.
However, target data from various environments is not always available during the training process. To overcome this limitation, research on domain generalization has recently gained attention, using only source data to train the model.

Synthetic to real domain generalization offers a possibility to exploit the advantage of the availability of synthetic data. In \cite{Chen2020}, the synthetically trained network is encouraged to maintain similar representations as the ImageNet pre-trained model. In other works, style-diversified samples \cite{Zhao2022} or web-crawled images \cite{Kim2021} are utilized for improving the representational consistency between synthetic and real-world for the sake of generalizable semantic segmentation. The model presented in \cite{Shiau2021} is trained on multiple source domains (synthetic and real) to generalize to unseen data domains. The variety of contents and styles from ImageNet is leveraged in \cite{Lee2022} to learn domain-generalized semantic features. In \cite{Choi2021}, an instance selective whitening loss is introduced to disentangle the domain-specific style and domain-invariant content to remove only the style information causing domain shift.

In contrast to domain adaptation and generalization, our method does not require target domain data or a great amount of source domain data for training, we only consider depth information for domain generalization. Moreover, we do not modify the training process of the semantic segmentation network, i.e., we are independent of the network due to modularity. For these reasons, the presented approaches cannot be considered as suitable baselines. 
%
%
\paragraph{False Negative Reduction in Semantic Segmentation} 
Reducing false negatives, i.e., obtaining a higher recall rate, is often achieved in semantic segmentation by modifying the loss function. In \cite{Xiang20190}, a higher recall rate for a real-time DNN is obtained by modifying the loss function, classifier and decision rule. A similar approach presented in \cite{Xiang20191} considers an importance-aware loss function to improve a network's reliability. To reduce false negative segments of minority classes, differences between the Bayes and the Maximum Likelihood decision rule are exploited introducing class priors that assign larger weight to underrepresented classes \cite{Chan2020}. Since minority classes are not necessarily hard to predict, leading to the prediction of many false positives, a hard-class mining loss is introduced in \cite{Tian2021} by redesigning the cross entropy loss to dynamically weight the loss for each class based on instantaneous recall. In \cite{Zhong2021}, false negative pixels in semi-supervised semantic segmentation are reduced by using the pixel-level $\ell_{2}$ loss and the pixel contrastive loss.

While the presented approaches modify the training process and/or the decision rule, we post-process only the output of the semantic segmentation network. 
For the first time, we present a false negative reduction approach which overcomes domain gaps using depth information. The only work \cite{Maag2021} which also uses depth heatmaps addressing the recall rate improvement works on video instance segmentation. 
%
%
\paragraph{Improving Segmentation using Depth Estimation} 
The predictions of semantic segmentation and depth estimation masks are improved in previous works using joint network architectures sharing information for both tasks \cite{Chen2019,Jiao2018}. Furthermore, approaches are introduced where information of one task enhance the prediction quality of the other task. The semantic segmentation task is improved in \cite{Hazirbas2016} by an encoder consisting of two network branches which extract features from depth and RGB images simultaneously. In \cite{Cao2017}, RGB-D data is also fed into a network that extracts both RGB and depth features in parallel for semantic segmentation prediction (and object detection). Contrary, a single shared encoder is used in \cite{Novosel2019} to enhance performance for a supervised task, here semantic segmentation, which obtains information of two self-supervised tasks (colorization and depth prediction) exploiting unlabeled data. In \cite{Jiang2018}, a semantic segmentation network is pre-trained for depth prediction to serve as a powerful proxy for learning visual representations. In addition to learning features from depth information, a student-teacher framework is considered in \cite{Hoyer2021} to select the most helpful samples to annotate for semantic segmentation. 

In comparison to the described methods modifying the network architecture, our foreground-background prediction runs independently and in parallel with semantic segmentation inference, and the aggregation serves as lightweight post-processing step. In particular, we cannot regard the presented approaches as suitable baselines since the domain generalization capability is not tested. However, these methods demonstrate that depth information can be used to enhance semantic segmentation.
%
%
%
\section{\uppercase{Method Description}}\label{sec:method}
Our method is composed of two parallel branches, i.e., the image segmentation and depth segmentation branch, see \autoref{fig:overview}. The outputs of both streams are aggregated to detect segments overlooked by the basic semantic segmentation network. As many false positive segments can be generated by the fusion, false positive pruning is applied in an additional post-processing step.
%
%
\subsection{Foreground-Background Segmentation}\label{sec:depth_seg}
In this section, we introduce our modified segmentation network for foreground-background segmentation. We assume that a depth estimation (and a semantic segmentation ground truth) is available for each input image. Our approach is modular and independent of the choice of the depth estimation (and the semantic segmentation) network. The basis for the modified network can be any standard semantic segmentation network. However, instead of feeding an RGB image into the network a depth estimation heatmap is used and the semantic space is composed of only two classes - foreground and background.

The binarization into foreground and background is adapted from the \emph{thing} and \emph{stuff} decomposition in the computer vision field like in panoptic segmentation \cite{Kirillov2019}. Using automated driving as example application, things are countable objects such as persons, animals, cars or bicycles. The stuff classes consist of amorphous regions of similar texture or material such as sky, road, nature or buildings. Note, the idea of things and stuff also exists in other application areas like robot navigation.
%
%
\subsection{Aggregation of Predictions}\label{sec:aggregation}
From the first branch, we obtain a semantic segmentation prediction, i.e., a pixel-wise classification of image content. The DNN provides for each pixel $z$ a probability distribution $f_{z}(y|x)$ over a prescribed label space $y \in \mathcal{C} = \{y_{1}, \ldots, y_{c} \}$ with $c$ different class labels, given an input image $x$. 
The predicted class for each pixel $z$ is computed by the maximum a-posteriori principle
\begin{equation}\label{eq:pred_class}
    \hat y_z(x)=\argmax_{y\in\mathcal{C}}f_z(y|x) \, .
\end{equation}
The second branch provides a foreground-background segmentation. Given the same input image $x$, we obtain for each pixel $z$ the probability of being a foreground pixel $g_z(x) \in [0,1]$ considering a binary classification problem.
The predicted segmentations are aggregated pixel-wise resulting in a combined prediction with the class label background $y_{0}$ or a foreground class label $y \in \tilde{\mathcal{C}} \subset \mathcal{C}$ per pixel. For this, we split the label space into foreground class labels $\tilde{\mathcal{C}} = \{y_{1}, \ldots, y_{\tilde{c}} \}$, $\tilde{c} < c$, and background class labels $\{y_{\tilde{c}+1}, \ldots, y_{c} \}$ with $y_{0} = \mathcal{C} \setminus \tilde{\mathcal{C}}$. 
The combination is defined per pixel by
\begin{equation}
\hat{s}_{z} (x) = \begin{cases} 
\hat y_z(x), &\text{ if } \hat y_z(x) \in \tilde{\mathcal{C}} \\
\displaystyle \argmax_{y\in\tilde{\mathcal{C}}}f_z(y|x), &\text{ if } g_z(x) > 0.5 \wedge \hat{y}_z(x) \notin \tilde{\mathcal{C}} \\
y_{0}, &\text{ else } \, .
\end{cases}
\end{equation}
If the semantic segmentation network predicts a foreground class or the foreground-background network predicts foreground, the pixel is considered as foreground and assigned to the foreground class $y \in \tilde{\mathcal{C}}$ of the semantic segmentation with the highest probability. Otherwise, the pixel is assigned to the class background.
Moreover, $\hat{\mathcal S}_x=\{ \hat{s}_{z} (x) | z\in x\}$ denotes the combined segmentation consisting of foreground classes and the background class. 
%
%
\subsection{Meta Classification}\label{sec:meta_classif}
The combination of the semantic segmentation and the foreground-background prediction can increase the number of false positives. For this reason, we apply meta classification \cite{Rottmann2018} as false positive pruning step using uncertainty measures. The degree of randomness in semantic segmentation prediction $f_{z}(y|x)$ is quantified by (pixel-wise) dispersion measures, like the entropy. 
To obtain segment-wise features characterizing uncertainty of a given segment from these pixel-wise dispersion measures, we aggregate them over segments by average pooling.
In addition, we hand-craft features based on object's geometry like the segment size or the geometric center obtaining uncertainty information.
These hand-crafted measures form a structured dataset where the rows correspond to predicted segments and the columns to features. A detailed description of these hand-crafted features can be found in \autoref{app:meta_classif}.

To determine if a predicted segment is a false positive, i.e., has no overlap with a ground truth segment of a foreground class, we consider the intersection over union ($\IoU$, \cite{Jaccard1912}), a typical performance measure of segmentation networks with respect to the ground truth.
Meta classification tackles the task of classifying between $\IoU = 0$ (false positive) and $\IoU > 0$ (true positive) for all predicted segments. If a segment is predicted to be a false positive, it is no longer considered as a foreground segment but as background. We perform meta classification using our structured dataset as input. Note, these hand-crafted measures are computed without the knowledge of the ground truth data. To train the classifier, we use gradient boosting \cite{Friedman2002} that outperforms linear models and shallow neural networks as shown in \cite{Maag2020}. We study to which extent our aggregated prediction followed by meta classification improves the detection performance for important classes compared to basic semantic segmentation.
%
%
%
\section{\uppercase{Experiments}}\label{sec:result}
In this section, we first present the experimental setting and then demonstrate the performance improvements of our method compared to the basic semantic segmentation network in terms of false negative reduction overcoming the domain gap.
%
%
\subsection{Experimental Setting}
%
%
\paragraph{Datasets}
We perform our tests on four datasets for semantic segmentation in street scenes considering Cityscapes \cite{Cordts2016} as source domain and A2D2 \cite{Geyer2020}, LostAndFound \cite{Pinggera2016} as well as India Driving (IDD) \cite{Varma2019} as target domains. The training/validation split of Cityscapes consists of $2,\!975$/$500$ images from dense urban traffic in $18$/$3$ different German towns, respectively. Thus, our foreground class consists of all road user classes, i.e., human (person and rider) and vehicle (car, truck, bus, train, motorcycle and bicycle) and the background of categories flat, construction, object, nature and sky.
From the A2D2 dataset, we sample $500$ images out of $23$ image sequences for our tests covering urban, highways and country roads in three cities. This variety of environments is not included in the Cityscapes dataset resulting in a domain shift in the background. 
The validation set of LostAndFound containing $1,\!203$ images is designed for detecting small obstacles on the road in front of the ego-car. This causes a foreground domain shift as these objects are not contained in the semantic space of Cityscapes. 
We use $538$ frames of the IDD dataset which contains unstructured environments of Indian roads inducing a domain shift in both, foreground and background. The latter is caused by, for example, the diversity of ambient conditions and ambiguous road boundaries. The foreground domain shift occurs as the IDD dataset consists of two more relevant foreground classes (animals and auto rickshaws) and the Cityscapes foreground objects differ significantly.
%
%
\paragraph{Networks}
We consider the state-of-the-art DeepLabv3+ network \cite{Chen2018} with WideResNet38 \cite{Wu2016} as backbone and the more lightweight (and thus weaker) DualGCNet \cite{Zhang2019} with ResNet50 \cite{He2016} backbone for semantic segmentation. Both DNNs are trained on the Cityscapes dataset achieving $\mathit{mean}$ $\IoU$ ($\mIoU$) values of $90.29\%$ for DeepLabv3+ and $79.68\%$ for DualGCNet on the Cityscapes validation set.
For depth estimation trained on the KITTI dataset \cite{Geiger2013}, we use the supervised depth estimation network BTS \cite{Lee2019} with DenseNet-161 \cite{Huang2017} backbone obtaining a \emph{relative absolute error} on the KITTI validation set of $0.090$ and the unsupervised Monodepth2 \cite{Godard2019} with ResNet18 backbone achieving $0.106$ \emph{relative absolute error}.
Our modified segmentation network is based on the DeepLabv3+ architecture with WideResNet38 backbone having high predictive power and is fed with depth estimation heatmaps of the Cityscapes dataset predicted by the BTS network and Monodepth2, respectively. We train this network on the training split of the Cityscapes dataset and use the binarized (into foreground and background) semantic segmentation ground truth to compare our results with the basic semantic segmentation network which is also trained on Cityscapes. For the BTS network a validation $\mIoU$ of $88.34\%$ is obtained and for Monodepth2 of $85.12\%$.
%
%
\paragraph{Evaluation Metrics}
Meta classification provides a probability of observing a false positive segment and such a predicted false positive segment is considered as background. We threshold on this probability with $101$ different values $h \in H = \{ 0.00,0.01, \ldots, 0.99, 1.00 \} $. For each threshold, we calculate the number of true positive, false positive and false negative foreground segments resulting in precision ($\mathit{prec}(h)$) and recall ($\mathit{rec}(h)$) values on segment-level depended of $h$. 
The degree of separability is then computed as the area under precision recall curve ($\auc$) by thresholding the meta classification probability. In addition, we compute the recall rate at $80\%$ precision rate ($\rec$) for the evaluation. Furthermore, we consider the segment-wise $F_{1}$ score which is defined by $F_{1} (h) = 2 \cdot \mathit{prec}(h) \cdot \mathit{rec}(h) / (\mathit{prec}(h) + \mathit{rec}(h))$. To obtain an evaluation metric independent of the meta classification threshold $h$, we calculate the averaged $F_{1}$ score $\bar{F_1} = \frac{1}{|H|} \sum_{h \in H} F_{1}(h)$ and the optimal $F_{1}$ score $F_1^* = \max_{h \in H} F_{1} (h)$. For a detailed description of these metrics see \autoref{app:metrics}.
%
%
\subsection{Numerical Results}
%
%
\paragraph{Results on the Source Domain}
First, we study the predictive power of the meta classifier trained on the Cityscapes (validation) dataset using a train/test splitting of $80\%/20\%$ shuffling $5$ times, such that all segments are a part of the test set. We use meta classification to prune possible false positive segments that are falsely predicted as foreground segments. For the comparison of basic semantic segmentation performance with our approach, meta classifiers are trained on the predicted foreground segments, respectively. These classifiers achieve test classification $\auroc$ values between $94.68\%$ and $99.14\%$. The $\auroc$ (area under receiver operating characteristic curve) is obtained by varying the decision threshold in a binary classification problem, here for the decision between $\IoU=0$ and $>0$. The influence of meta classification on the performance is studied in \autoref{app:effect_meta_classif}.

\begin{figure*}[t]
    \center
    \includegraphics[width=0.97\textwidth]{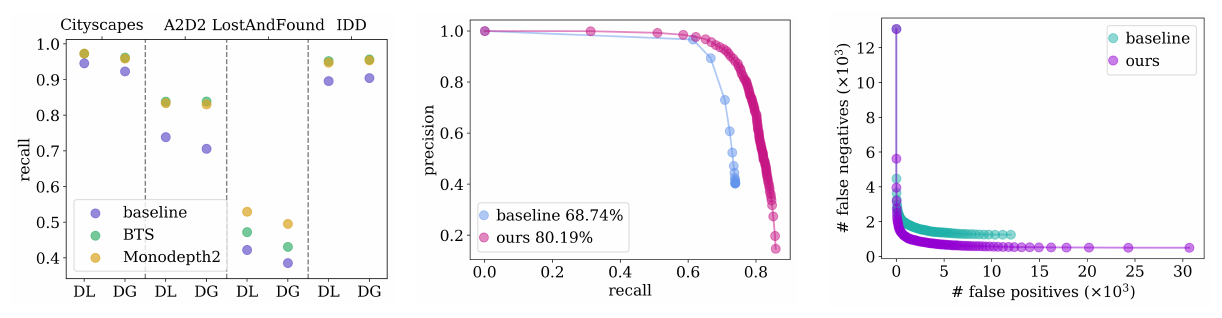}
    \caption{\emph{Left}: The recall values under the assumption of same precision values for all datasets and networks. We distinguish the performance for the DeepLabv3+ (DL) and the DualGCNet (DG) semantic segmentation networks whose predictions serve as baselines. We compare these with our approach using the BTS and the Monodepth2 depth estimation network, respectively. \emph{Center}: Precision-recall curves for the A2D2 dataset, the DeepLabv3+ and BTS networks. \emph{Right}: Number of false positive vs.\ false negative segments for different meta classification thresholds for the IDD dataset, the DualGCNet and BTS networks using $20\%$ of this dataset for fine-tuning.}
    \label{fig:curves}
\end{figure*}

We compare the detection performances which are shown in \autoref{tab:auprc} using presented evaluation metrics.
\begin{table}[t]
\caption{Performance results for the Cityscapes dataset for the basic semantic segmentation prediction vs.\ our approach, i.e., the DeepLabv3+/DualGCNet prediction aggregated with foreground-background prediction using BTS or Monodepth2.}
\centering
\scalebox{0.75}{
\begin{tabular}{c|cccc}
\cline{1-5}
\multicolumn{1}{c}{} & $\auc$ & $\bar{F_1}$ & $F_1^*$ & $\rec$ \rule{0mm}{3.4mm}\\
\cline{1-5}
DeepLabv3+ & $94.26$ & $\mathbf{90.61}$ & $94.69$ & $94.49$ \\
+ BTS & $97.07$ & $90.21$ & $\mathbf{95.80}$ & $\mathbf{97.15}$ \\
+ Monodepth2 & $\mathbf{97.08}$ & $90.03$ & $95.73$ & $\mathbf{97.15}$ \\
\cline{1-5}
DualGCNet & $91.85$ & $\mathbf{88.68}$ & $92.77$ & $92.18$ \\
+ BTS & $\mathbf{95.90}$ & $87.92$ & $\mathbf{94.66}$ & $\mathbf{95.88}$ \\
+ Monodepth2 & $95.63$ & $87.94$ & $94.58$ & $95.66$ \\
\cline{1-5}
\end{tabular} }
\label{tab:auprc}
\end{table}
We observe that our method obtain higher $\auc$, $F_1^*$ and $\rec$ values than the semantic segmentation prediction.
Note, there is no consistency on which depth estimation network yields more enhancement.
In particular, we reduce the number of non-detected segments of foreground classes. In \autoref{fig:curves} (left), the highest recall values of the semantic segmentation predictions are shown, i.e., no segments are deleted using meta classification.
For our method, we use the meta classification threshold where the precision of our method is equal to that of the baseline. As a consequence, for the identical precision values we obverse an increase in recall by up to $2.71$ percent points (pp) for the Cityscapes dataset. In \autoref{app:results_class}, 
more numerical results evaluated on individual foreground classes are presented.

\begin{figure*}[t]
    \center
    \includegraphics[width=0.99\textwidth]{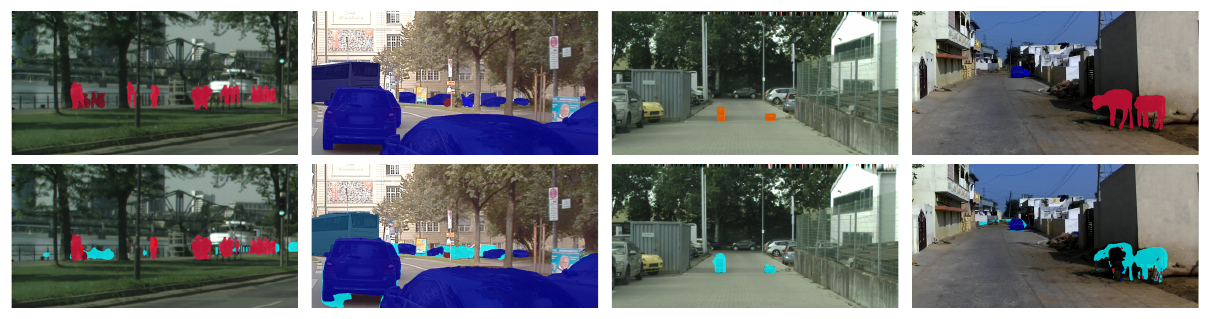}
    \caption{Examples for segments that are overlooked by the basic semantic segmentation network and detected by our approach for Cityscapes (DualGCNet, BTS, \emph{left}), A2D2 (DeepLabv3+, BTS, \emph{center left}), LostAndFound (DeepLabv3+, Monodepth2, \emph{center right}) and IDD dataset (DeepLabv3+, BTS, \emph{right}). \emph{Top}: Ground truth images including only the labels of foreground classes. \emph{Bottom}: Basic semantic segmentation prediction in typical Cityscapes colors for foreground segments (shades of blue and red) as well as the foreground prediction of our modified segmentation network (cyan).}
    \label{fig:examples}
\end{figure*}

The $\mIoU$ is the commonly used performance measure for semantic segmentation. To compute the $\mIoU$ for the aggregated prediction $\hat{\mathcal S}_x$, we have to fill the background values as they are ignored up to now. Similar to how we obtain the foreground class during the combination, we assign to every background pixel the background class $y \in \mathcal{C} \setminus \tilde{\mathcal{C}}$ of the semantic segmentation with the highest probability. The results for semantic segmentation prediction and the difference to our aggregated predictions are shown in the Cityscapes column of \autoref{tab:miou}.
\begin{table}[t]
\caption{$\mIoU$ results for both semantic segmentation networks and the difference to our approach. A higher $\mIoU$ value corresponds to better performance.}
\centering
\scalebox{0.75}{
\begin{tabular}{c|ccc}
\cline{1-4}
\multicolumn{1}{c}{} & Cityscapes & A2D2 & IDD \rule{0mm}{2.5mm}\\
\cline{1-4}
DeepLabv3+ & $90.29\%$ & $61.98\%$ & $57.26\%$ \\
+ BTS & $-4.72$ pp & $-2.87$ pp & $-1.59$ pp \\
+ Monodepth2 & $-5.97$ pp & $-4.84$ pp & $-3.99$ pp \\
\cline{1-4}
DualGCNet & $79.68\%$ & $23.76\%$ & $45.79\%$ \\
+ BTS & $-3.99$ pp & $+0.12$ pp & $-1.03$ pp \\
+ Monodepth2 & $-5.09$ pp & $-0.49$ pp & $-3.12$ pp \\
\cline{1-4}
\end{tabular} }
\label{tab:miou}
\end{table}
We perform slightly worse in the overall performance accuracy ($\mIoU$) as the foreground-background masks are location-wise less accurate than the segmentation masks, see \autoref{fig:examples}. The reason is that the modified segmentation network is fed with predicted depth heatmaps which may be inaccurate resulting in less precise separation of foreground and background.
Nonetheless, we detect foreground objects, here road users, that are overlooked by the semantic segmentation network (for example, see the bicycle in \autoref{fig:examples} (left)). 
%
%
\paragraph{Results under Domain Shift}
In this section, we study the false negative reduction for the A2D2, LostAndFound and IDD datasets under domain shift from the source domain Cityscapes. 
As mentioned above, since the semantic segmentation networks as well as the modified segmentation networks are trained on the Cityscapes dataset, we train also the meta classification model on this dataset using all predicted segments. We obtain meta classification test $\auroc$ values up to $93.12\%$ for A2D2, $91.65\%$ for LostAndFound and $93.97\%$ for IDD.

We compare the performance of our approach with the semantic segmentation prediction by computing the evaluation metrics, results are given in \autoref{tab:auprc_domain}.
\begin{table*}[t]
\caption{Performance results for the basic semantic segmentation prediction vs.\ our approach.} 
\centering
\scalebox{0.72}{
\begin{tabular}{c|cccc|cccc|cccc}
\cline{1-13}
\multicolumn{1}{c}{} & \multicolumn{4}{c}{A2D2} & \multicolumn{4}{c}{LostAndFound} & \multicolumn{4}{c}{IDD} \rule{0mm}{2.5mm}\\
\cline{1-13}
\multicolumn{1}{c}{} & $\auc$ & $\bar{F_1}$ & $F_1^*$ & $\rec$ & $\auc$ & $\bar{F_1}$ & $F_1^*$ & $\rec$ & $\auc$ & $\bar{F_1}$ & $F_1^*$ & $\rec$ \rule{0mm}{3.4mm}\\
\cline{1-13}
DeepLabv3+ & $68.74$ & $52.72$ & $76.36$ & $70.80$ & $40.05$ & $50.27$ & $53.06$ & $39.40$ & $84.11$ & $69.19$ & $87.79$ & $88.52$ \\
+ BTS & $\mathbf{80.19}$ & $\mathbf{66.96}$ & $80.46$ & $\mathbf{77.17}$ & $46.18$ & $51.08$ & $57.80$ & $45.06$ & $\mathbf{93.86}$ & $\mathbf{78.48}$ & $\mathbf{91.75}$ & $\mathbf{93.26}$ \\
+ Monodepth2 & $80.01$ & $66.77$ & $\mathbf{80.72}$ & $77.15$ & $\mathbf{51.67}$ & $\mathbf{54.15}$ & $\mathbf{60.41}$ & $\mathbf{48.07}$ & $93.35$ & $76.69$ & $91.45$ & $92.86$ \\
\cline{1-13}
DualGCNet & $48.64$ & $27.93$ & $\mathbf{65.48}$ & $\mathbf{58.67}$ & $36.80$ & $45.85$ & $50.21$ & $36.08$ & $84.59$ & $66.40$ & $87.37$ & $88.20$ \\
+ BTS & $42.16$ & $36.27$ & $51.22$ & $26.90$ & $42.34$ & $47.27$ & $53.59$ & $40.12$ & $92.23$ & $71.85$ & $89.78$ & $\mathbf{92.53}$ \\
+ Monodepth2 & $\mathbf{49.03}$ & $\mathbf{39.01}$ & $57.26$ & $19.29$ & $\mathbf{47.92}$ & $\mathbf{49.11}$ & $\mathbf{57.06}$ & $\mathbf{44.16}$ & $\mathbf{92.39}$ & $\mathbf{73.17}$ & $\mathbf{89.82}$ & $92.23$ \\
\cline{1-13}
\end{tabular} }
\label{tab:auprc_domain}
\end{table*}
\begin{table*}[t]
\caption{Evaluation results obtained by different splittings that are used for fine-tuning the meta classifier.}
\centering
\scalebox{0.72}{
\begin{tabular}{c|c|cccc|cccc|cccc}
\cline{1-14}
\multicolumn{2}{c}{} & \multicolumn{4}{c}{A2D2} & \multicolumn{4}{c}{LostAndFound} & \multicolumn{4}{c}{IDD} \rule{0mm}{2.5mm}\\
\cline{1-14}
\multicolumn{2}{c}{} & $\auc$ & $\bar{F_1}$ & $F_1^*$ & $\rec$ & $\auc$ & $\bar{F_1}$ & $F_1^*$ & $\rec$ & $\auc$ & $\bar{F_1}$ & $F_1^*$ & $\rec$ \rule{0mm}{3.4mm}\\
\cline{1-14}
             & $0\%$ & $80.19$ & $66.96$ & $80.46$ & $77.17$ & $46.18$ & $51.08$ & $57.80$ & $45.06$ & $93.86$ & $78.48$ & $91.75$ & $93.26$ \\
DeepLabv3+   & $20\%$ & $83.65$ & $75.90$ & $85.13$ & $82.01$ & $48.79$ & $60.30$ & $63.57$ & $48.92$ & $94.65$ & $82.12$ & $93.24$ & $93.85$ \\
+            & $40\%$ & $83.72$ & $\mathbf{76.03}$ & $85.25$ & $82.25$ & $49.01$ & $60.89$ & $64.01$ & $49.04$ & $94.66$ & $\mathbf{82.24}$ & $\mathbf{93.46}$ & $93.79$ \\
BTS          & $60\%$ & $\mathbf{83.75}$ & $75.99$ & $85.39$ & $82.05$ & $\mathbf{49.11}$ & $\mathbf{61.41}$ & $\mathbf{64.67}$ & $49.16$ & $\mathbf{94.86}$ & $82.19$ & $93.43$ & $93.89$ \\
             & $80\%$ & $83.67$ & $75.89$ & $\mathbf{85.50}$ & $\mathbf{82.27}$ & $48.88$ & $61.33$ & $64.53$ & $\mathbf{49.28}$ & $94.67$ & $82.23$ & $93.42$ & $\mathbf{93.93}$ \\
\cline{1-14}
             & $0\%$ & $80.01$ & $66.77$ & $80.72$ & $77.15$ & $51.67$ & $54.15$ & $60.41$ & $48.07$ & $93.35$ & $76.69$ & $91.45$ & $92.86$ \\
DeepLabv3+   & $20\%$ & $82.91$ & $76.00$ & $85.02$ & $81.79$ & $55.77$ & $63.70$ & $68.58$ & $55.72$ & $94.16$ & $81.24$ & $92.58$ & $93.48$ \\
+            & $40\%$ & $\mathbf{83.12}$ & $\mathbf{76.23}$ & $84.98$ & $\mathbf{82.18}$ & $\mathbf{56.19}$ & $64.44$ & $69.25$ & $56.14$ & $\mathbf{94.21}$ & $81.36$ & $\mathbf{92.69}$ & $\mathbf{93.59}$ \\
Monodepth2   & $60\%$ & $83.10$ & $76.11$ & $\mathbf{85.19}$ & $81.96$ & $56.06$ & $\mathbf{64.57}$ & $69.58$ & $56.20$ & $94.19$ & $\mathbf{81.41}$ & $92.65$ & $93.47$ \\
             & $80\%$ & $83.03$ & $75.94$ & $85.15$ & $81.68$ & $56.11$ & $64.50$ & $\mathbf{69.65}$ & $\mathbf{56.39}$ & $94.17$ & $81.31$ & $92.50$ & $93.49$ \\
\cline{1-14}
             & $0\%$ & $42.16$ & $36.27$ & $51.22$ & $26.90$ & $42.34$ & $47.27$ & $53.59$ & $40.12$ & $92.23$ & $71.85$ & $89.78$ & $92.53$ \\
DualGCNet    & $20\%$ & $82.59$ & $74.59$ & $83.06$ & $80.56$ & $45.76$ & $56.88$ & $60.12$ & $45.48$ & $94.71$ & $81.85$ & $92.56$ & $\mathbf{93.79}$ \\
+            & $40\%$ & $\mathbf{82.85}$ & $\mathbf{74.83}$ & $\mathbf{83.82}$ & $\mathbf{81.26}$ & $46.15$ & $58.08$ & $61.24$ & $46.14$ & $\mathbf{94.74}$ & $\mathbf{81.99}$ & $\mathbf{92.75}$ & $93.70$ \\
BTS          & $60\%$ & $82.82$ & $74.72$ & $83.58$ & $81.20$ & $46.19$ & $58.15$ & $61.26$ & $\mathbf{46.39}$ & $94.71$ & $81.85$ & $92.73$ & $93.77$ \\
             & $80\%$ & $82.76$ & $74.53$ & $83.53$ & $81.20$ & $\mathbf{46.24}$ & $\mathbf{58.53}$ & $\mathbf{61.82}$ & $46.27$ & $94.67$ & $81.85$ & $92.65$ & $93.68$ \\
\cline{1-14}
             & $0\%$ & $49.03$ & $39.01$ & $57.26$ & $19.29$ & $47.92$ & $49.11$ & $57.06$ & $44.16$ & $92.39$ & $73.17$ & $89.82$ & $92.23$ \\
DualGCNet    & $20\%$ & $81.98$ & $74.66$ & $83.26$ & $80.69$ & $53.14$ & $61.26$ & $65.97$ & $53.01$ & $94.27$ & $81.56$ & $91.93$ & $93.31$ \\
+            & $40\%$ & $82.27$ & $74.97$ & $83.28$ & $81.02$ & $53.77$ & $62.30$ & $66.95$ & $53.61$ & $94.39$ & $81.60$ & $92.02$ & $\mathbf{93.51}$ \\
Monodepth2   & $60\%$ & $\mathbf{82.31}$ & $\mathbf{75.14}$ & $\mathbf{83.34}$ & $\mathbf{81.35}$ & $\mathbf{53.91}$ & $62.48$ & $67.17$ & $\mathbf{53.80}$ & $\mathbf{94.40}$ & $\mathbf{81.64}$ & $\mathbf{92.32}$ & $93.31$ \\
             & $80\%$ & $82.06$ & $74.42$ & $82.91$ & $80.82$ & $53.76$ & $\mathbf{62.59}$ & $\mathbf{67.21}$ & $53.73$ & $94.34$ & $81.54$ & $91.96$ & $93.47$ \\
\cline{1-14}
\end{tabular} }
\label{tab:study_classif}
\end{table*}
The performance metrics are greatly increased by our method demonstrating that our approach is more robust to domain shift. Noteworthy, we outperform the stronger DeepLabv3+ network in all cases. Example curves are presented in \autoref{fig:curves} (center) for the A2D2 dataset where an $\auc$ enhancement of $11.45$ pp is obtained. Our precision-recall curve is entirely above the baseline. In particular, for identical precision values, we obtain an increase in recall by up to $13.24$ pp, i.e., reduce the number of false negative segments, as also shown in \autoref{fig:curves} (left). Examples for detected segments that are missed by the semantic segmentation network are given in \autoref{fig:examples} for all datasets. Hence, our method detect segments of well-trained classes, i.e., the overlooked bicycle in the Cityscapes dataset or various cars in A2D2. Moreover, we bridge the domain gap as we find small obstacles (LostAndFound) and animals (IDD) that are not part of the Cityscapes dataset and thus, are not included in the semantic space for training. In \autoref{app:results_class}, 
more numerical results evaluated on individual foreground classes are presented.

In \autoref{tab:miou}, the differences between the $\mIoU$ values are evaluated on the Cityscapes classes. For the A2D2 dataset, the classes are mapped to the Cityscapes ones and for the IDD dataset, we treat the additional classes animal as human and auto rickshaw as car. For the LostAndFound, an evaluation is not possible as it contains only labels for the road and the small obstacles which do not fit into the semantic space. With one positive exception, we are slightly worse in overall accuracy performance. On the one hand, the images in \autoref{fig:examples} demonstrate why we decrease the accuracy slightly as the predictions and in particular, the segment boundaries are less accurate. On the other hand, these images motivate the benefit of our method as completely overlooked segments are detected. Furthermore, we bridge the domain shift in a post-processing manner that only requires two more inferences which run in parallel to semantic segmentation prediction. 
%
%
\paragraph{Fine-Tuning of the Meta Classifier}
Up to now, we have trained the segmentation networks as well as the meta classifier on Cityscapes for the experiments on A2D2, LostAndFound and IDD dataset. In this paragraph, we investigate the predictive power of the meta classifier and the implications on false negative reduction using parts of the target dataset for fine-tuning. Note, this domain adaptation only occurs in the post-processing meta classification step (retraining the neural networks is not necessary) and thus, the fine-tuning is lightweight and requires only a small amount of ground truth data. In detail, we retrain the meta classifier with $20\%$, $40\%$, $60\%$ and $80\%$ of the target dataset, respectively. The corresponding performance results are shown in \autoref{tab:study_classif}. 
We observe great enhancements even with only a fine-tuning of $20\%$ of the target domain obtaining an increase of up to $40.43$ pp for $\auc$. The maximal increase is achieved for the A2D2 dataset (on the DualGCNet and BTS networks) for which $20\%$ correspond to about $100$ images that are used for retraining and achieving such an improvement. For all datasets, the greatest performance gap occurs between a trained meta classifier only on the Cityscapes dataset and using a small amount of the target domain data (here $20\%$). Increasing the subset of the target data, the performance is only slightly enhanced. Using $20\%$ for fine-tuning, the highest $\auc$ value of $94.71\%$ is obtained by the DualGCNet and the BTS network on the IDD dataset. The corresponding number of false positives and false negatives is given in \autoref{fig:curves} (right). Note, the meta classifier for the baseline prediction is trained on the same train splitting. We outperform the basic semantic segmentation prediction and thus achieve a lower number of detection errors, in particular false negatives, therefore bridging the domain gap.
%
%
\section{\uppercase{Conclusion and Outlook}}\label{sec:conclusion} 
In this work, we proposed a domain generalization method applicable to any semantic segmentation network using monocular depth estimation, in particular reducing non-detected segments. We inferred a depth heatmap via a modified segmentation network that predicts foreground-background masks in parallel to a semantic segmentation network. Aggregating both predictions in an uncertainty-aware manner with a focus on important classes, false negative segments were successfully reduced. Our experiments suggest that also in a single-sensor setup, the information about spatial structure from pre-trained monocular depth estimators can be utilized well to improve the robustness of off-the-shelf segmentation networks under domain shift in various settings.

As an extension of this work, it might be interesting to apply our method on other application fields like robotic navigation. In this case, a differentiation between foreground and background classes is possible, for example, the items in robotic navigation are most important. Our approach is applicable without major modifications, since trained depth estimation networks are available for several tasks and our modified segmentation network can be retrained on the same data like the basic semantic segmentation network uses. If a task-specific segmentation network is required, then as foreground-background network the semantic segmentation network architecture can be used only with small modifications on input and output.
%
%
\section*{\uppercase{Acknowledgements}}
We thank M.\! K.\! Neugebauer for support in data handling and programming. This work is supported by the Ministry of Culture and Science of the German state of North Rhine-Westphalia as part of the KI-Starter research funding program.

\bibliographystyle{apalike}
{\small
\bibliography{egbibs}}

\appendix
\section*{\uppercase{Appendix}}
\begin{table*}[t]
\caption{Evaluation results using meta classification ($F_1^*$) and without ($F_1(1)$) for the basic semantic segmentation prediction (DeepLabv3+/DualGCNet) and our approach, i.e., the DeepLabv3+/DualGCNet prediction aggregated with foreground-background prediction using BTS or Monodepth2.} 
\centering
\scalebox{0.82}{
\begin{tabular}{c|cc|cc|cc|cc}
\cline{1-9}
\multicolumn{1}{c}{} & \multicolumn{2}{c}{Cityscapes} & \multicolumn{2}{c}{A2D2} & \multicolumn{2}{c}{LostAndFound} & \multicolumn{2}{c}{IDD} \rule{0mm}{2.5mm}\\
\cline{1-9}
\multicolumn{1}{c}{} & $F_1(1)$ & $F_1^*$ & $F_1(1)$ & $F_1^*$ & $F_1(1)$ & $F_1^*$ & $F_1(1)$ & $F_1^*$ \rule{0mm}{2.8mm}\\
\cline{1-9}
DeepLabv3+ & $\mathbf{84.00}$ & $94.69$ & $\mathbf{52.16}$ & $76.36$ & $\mathbf{49.54}$ & $53.06$ & $\mathbf{69.14}$ & $87.79$ \\
+ BTS & $43.16$ & $\mathbf{95.80}$ & $25.09$ & $80.46$ & $40.19$ & $57.80$ & $39.00$ & $\mathbf{91.75}$  \\
+ Monodepth2 & $38.11$ & $95.73$ & $17.16$ & $\mathbf{80.72}$ & $33.61$ & $\mathbf{60.41}$ & $25.73$ & $91.45$ \\
\cline{1-9}
DualGCNet & $\mathbf{82.82}$ & $92.77$ & $\mathbf{25.89}$ & $\mathbf{65.48}$ & $\mathbf{45.88}$ & $50.21$ & $\mathbf{64.11}$ & $87.37$ \\
+ BTS & $53.99$ & $\mathbf{94.66}$ & $25.40$ & $51.22$ & $40.17$ & $53.59$ & $44.64$ & $89.78$ \\
+ Monodepth2 & $50.92$ & $94.58$ & $23.98$ & $57.26$ & $35.07$ & $\mathbf{57.06}$ & $35.87$ & $\mathbf{89.82}$ \\
\cline{1-9}
\end{tabular} }
\label{tab:meta_improve}
\end{table*}
\begin{table*}[t]
\caption{Performance results for the basic semantic segmentation prediction (DeepLabv3+/DualGCNet) vs.\ our approach, i.e., the DeepLabv3+/DualGCNet prediction aggregated with foreground-background prediction using BTS or Monodepth2, for class person, car and bicycle.}
\centering
\scalebox{0.82}{
\begin{tabular}{c|c|ccc|ccc|ccc}
\cline{1-11}
\multicolumn{2}{c}{} & \multicolumn{3}{c}{Cityscapes} & \multicolumn{3}{c}{A2D2} & \multicolumn{3}{c}{IDD} \rule{0mm}{2.5mm}\\
\cline{1-11}
 \multicolumn{2}{c}{} & $\auc$ & $\bar{F_1}$ & $F_1^*$ & $\auc$ & $\bar{F_1}$ & $F_1^*$ & $\auc$ & $\bar{F_1}$ & $F_1^*$ \rule{0mm}{3.4mm}\\
\cline{1-11}
person & DeepLabv3+   & $83.11$ & $80.33$ & $84.66$ & $40.66$ & $40.19$ & $54.83$ & $39.70$ & $35.31$ & $56.78$ \\
 & + BTS        & $\mathbf{87.36}$ & $\mathbf{80.36}$ & $\mathbf{86.89}$ & $47.10$ & $\mathbf{43.37}$ & $55.43$ & $46.14$ & $\mathbf{41.05}$ & $54.94$ \\
 & + Monodepth2 & $86.87$ & $80.27$ & $86.62$ & $\mathbf{48.85}$ & $41.83$ & $\mathbf{56.60}$ & $\mathbf{48.88}$ & $40.73$ & $\mathbf{56.80}$ \\
\cline{2-11}
& DualGCNet    & $75.05$ & $\mathbf{73.30}$ & $77.12$ & $\mathbf{13.67}$ & $17.64$ & $\mathbf{35.34}$ & $31.29$ & $25.37$ & $47.28$ \\
 & + BTS        & $\mathbf{79.72}$ & $73.24$ & $\mathbf{80.30}$ & $13.28$ & $20.68$ & $28.42$ & $36.94$ & $30.23$ & $47.72$ \\
 & + Monodepth2 & $78.39$ & $72.78$ & $79.70$ & $13.56$ & $\mathbf{21.28}$ & $28.61$ & $\mathbf{41.51}$ & $\mathbf{31.48}$ & $\mathbf{48.08}$ \\
\cline{1-11}
car & DeepLabv3+   & $86.19$ & $85.18$ & $89.31$ & $64.77$ & $56.21$ & $73.16$ & $55.25$ & $39.92$ & $70.03$ \\
 & + BTS        & $\mathbf{89.20}$ & $\mathbf{85.69}$ & $\mathbf{90.61}$ & $\mathbf{75.53}$ & $66.74$ & $77.44$ & $\mathbf{69.99}$ & $\mathbf{50.97}$ & $\mathbf{73.45}$ \\
 & + Monodepth2 & $88.83$ & $85.15$ & $90.20$ & $74.82$ & $\mathbf{67.26}$ & $\mathbf{77.45}$ & $68.03$ & $49.70$ & $72.59$ \\
\cline{2-11}
 & DualGCNet    & $81.76$ & $\mathbf{81.45}$ & $85.44$ & $39.65$ & $21.93$ & $\mathbf{59.85}$ & $57.28$ & $41.89$ & $69.07$ \\
 & + BTS        & $\mathbf{85.88}$ & $81.31$ & $\mathbf{87.39}$ & $33.52$ & $30.37$ & $46.14$ & $63.12$ & $46.67$ & $69.70$ \\
 & + Monodepth2 & $85.24$ & $81.24$ & $87.13$ & $\mathbf{40.63}$ & $\mathbf{34.00}$ & $53.89$ & $\mathbf{64.52}$ & $\mathbf{48.50}$ & $\mathbf{69.93}$ \\
\cline{1-11}
bicycle & DeepLabv3+   & $85.46$ & $\mathbf{80.25}$ & $\mathbf{87.43}$ & $37.37$ & $47.00$ & $49.67$ & $14.36$ & $9.73$ & $33.73$ \\
 & + BTS        & $86.99$ & $79.02$ & $86.64$ & $\mathbf{43.78}$ & $\mathbf{48.47}$ & $\mathbf{54.88}$ & $21.70$ & $13.70$ & $38.25$ \\
 & + Monodepth2 & $\mathbf{87.20}$ & $78.54$ & $86.93$ & $42.37$ & $46.97$ & $53.24$ & $\mathbf{23.78}$ & $\mathbf{13.79}$ & $\mathbf{39.84}$ \\
\cline{2-11}
 & DualGCNet    & $77.05$ & $\mathbf{73.43}$ & $80.62$ & $\mathbf{15.22}$ & $19.85$ & $\mathbf{27.27}$ & $16.06$ & $8.68$ & $\mathbf{36.79}$ \\
 & + BTS        & $79.49$ & $72.68$ & $81.16$ & $8.08$ & $\mathbf{21.30}$ & $23.08$ & $16.00$ & $9.56$ & $33.51$ \\
 & + Monodepth2 & $\mathbf{79.99}$ & $72.82$ & $\mathbf{81.93}$ & $10.23$ & $19.70$ & $21.28$ & $\mathbf{21.93}$ & $\mathbf{10.70}$ & $36.36$ \\
\cline{1-11}
\end{tabular} }
\label{tab:auprc_class}
\end{table*}
%
%
\section{Details on Meta Classification}\label{app:meta_classif}
\begin{figure}[t]
    \center
    \includegraphics[trim=550 302 0 60,clip,width=0.23\textwidth]{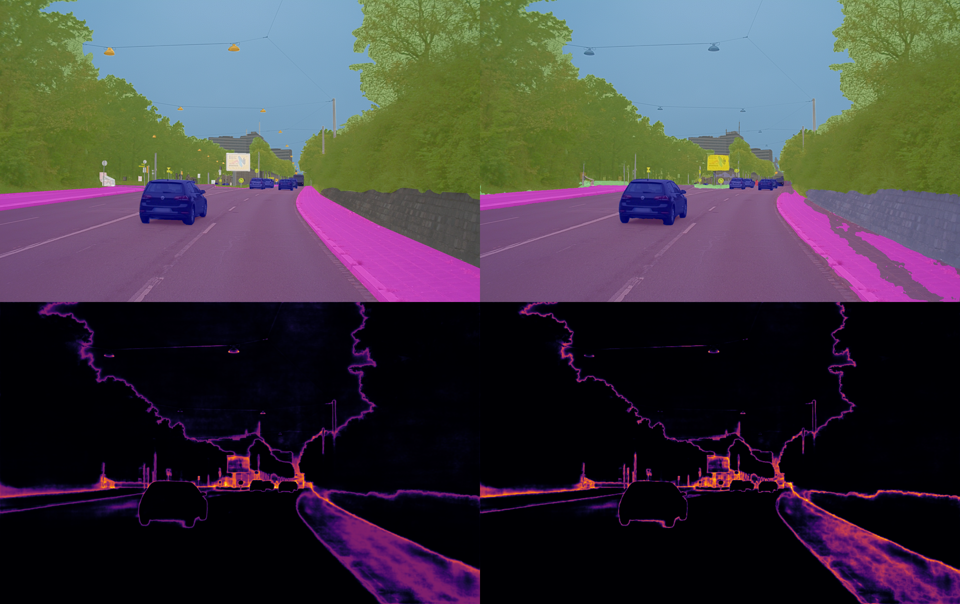}
    \includegraphics[trim=70 0 480 362,clip,width=0.23\textwidth]{figs/a2d2_frac_entro.png}
    \caption{\emph{Left}: Semantic segmentation predicted by a DNN. \emph{Right}: Entropy heatmap.}
    \label{fig:a2d2_entro}
\end{figure}

The semantic segmentation neural network provides for each pixel $z$ a probability distribution $f_{z}(y|x)$ over a label space $\mathcal{C} = \{y_{1}, \ldots, y_{c} \}$, with $y \in \mathcal{C}$ and given an input image $x$.
The degree of randomness in semantic segmentation prediction is quantified by (pixel-wise) dispersion measures, such as the entropy
\begin{equation}
    E_z(x) =-\frac{1}{\log(c)}\sum_{y\in \mathcal{C}}f_z(y|x)\log f_z(y|x) \, ,
\end{equation} (see \autoref{fig:a2d2_entro} (right))
the variation ratio 
\begin{equation}
    V_{z} = 1 - f_z(\hat y_z(x)|x) 
\end{equation}
or the probability margin 
\begin{equation}
    M_z(x) = V_{z} + \max_{y\in\mathcal{C}\setminus\{\hat y_z(x)\}} f_z(y|x) 
\end{equation}
with predicted class $\hat y_z(x)$ (see \autoref{eq:pred_class}).
Based on the different behavior of these measures and the segment's geometry for correct and false predictions, we construct segment-wise features by hand to quantify the observations that we made.
Let $\hat{\mathcal P}_x$ denote the set of predicted segments, i.e., connected components, (of the foreground class).
By aggregating these pixel-wise measures, segment-wise features are obtained and serve as input for the meta classifier.
To this end, we compute for each segment $q \in \hat{\mathcal P}_x$ the mean of the pixel-wise uncertainty values of a given segment, i.e., \emph{mean dispersions} $\bar D$, $D \in \{E,V,M\}$. Furthermore, we distinguish between the \emph{inner} of the segment $q_{in}\subset q$ consisting of all pixels whose eight neighboring pixels are also elements of $q$ and the \emph{boundary} $q_{bd}=  q \setminus q_{in}$. We observe that poor or false predictions are often accompanied by fractal segment shapes (a relatively large amount of boundary pixels). An example is shown in \autoref{fig:a2d2_entro} (left). This results in \emph{segment size} $S=|q|$ and mean dispersion features per segment also for the inner and the boundary since uncertainties may be higher on a segment's boundary (see \autoref{fig:a2d2_entro} (right)). Additionally, we define \emph{relative segment sizes} $\tilde S = S/S_{bd}$ and $\tilde S_{in} = S_{in}/S_{bd}$ quantifying the degree of fractality as well as \emph{relative mean dispersions} $\tilde {\bar D} = \bar D \tilde S$ and $\tilde {\bar D}_{in} = \bar D_{in} \tilde S_{in}$ where $D \in \{E,V,M\}$. 

For the foreground-background segmentation, given the same input image $x$, we obtain for each pixel $z$ the probability of being a foreground pixel $g_z(x) \in [0,1]$. Thus, we calculate the mean and relative entropy features for the foreground-background prediction (having only two classes), denoted by $\bar F_{*}$, $* \in \{\_,in,bd\}$, $\tilde {\bar F}$ and $\tilde {\bar F}_{in}$. Last, we add the \emph{geometric center}
\begin{equation}
\bar{q} = \frac{1}{S} \sum_{(z_{v}, z_{h}) \in q} (z_{v}, z_{h})
\end{equation}
where $(z_{v}, z_{h})$ describes the vertical and horizontal coordinate of pixel $z$ and the \emph{mean class probabilities} $P(y|q)$ for each foreground class $y \in \tilde{\mathcal{C}} \subset \mathcal{C}$ where $\tilde{\mathcal{C}} = \{y_{1}, \ldots, y_{\tilde{c}} \}$, $\tilde{c} < c$, to our set of hand-crafted features. 

Analogously to the set of predicted segments $\hat{\mathcal P}_x$, we denote by $\mathcal P_x$ the set of segments in the ground truth $\mathcal S_x$. To determine if a predicted segment $q \in \hat{\mathcal P}_x$ is a false positive, we consider the intersection over union. The segment-wise $\IoU$ is then defined as 
\begin{equation}
    \IoU (q) = \frac{| q \cap Q |}{| q \cup Q |}, \quad Q = \bigcup_{q' \in \mathcal P_x, q' \cap q \neq \emptyset} q' \, .
\end{equation}
%
%
\section{More Details on Evaluation Metrics}\label{app:metrics}
Let $\hat{\mathcal P}_x$ denote the set of predicted segments and $\mathcal P_x$ of ground truth segments. Meta classification provides a probability $m(q) \in [0,1]$ for each segment $q \in \hat{\mathcal P}_x$ to be a false positive on which we threshold with different values $h \in H = \{ 0.00,0.01, \ldots, 0.99, 1.00 \}$. A predicted false positive segment is considered as background. For each threshold $h$, we calculate over of all foreground segments in a given validation set $\mathcal{X}$ the number of false positives
\begin{equation}
    \mathrm{FP}(h) = \sum_{x \in \mathcal{X}} \sum_{ q \in \hat{\mathcal P}_x } 1_{ \{ \IoU (q) = 0 \} } \hspace{0.2ex} 1_{ \{ m(q) \leq h \} } \, ,
\end{equation}
true positives 
\begin{equation}
    \mathrm{TP}(h) = \sum_{x \in \mathcal{X}} \sum_{ q' \in \mathcal P_x} 1_{ \{ \IoU' (q,h) > 0 \} } 
\end{equation}
and false negatives 
\begin{equation}
    \mathrm{FN}(h) = \sum_{x \in \mathcal{X}} \sum_{ q' \in \mathcal P_x} 1_{ \{ \IoU' (q,h) = 0 \} } 
\end{equation}
where the indicator function is defined as
\begin{equation}
1_{ \{ A \} } = \begin{cases} 
1, &\text{ if } \text{event } A \text{ happens} \\
0, &\text{ else } \, 
\end{cases}
\end{equation}
and the $\IoU$ for a ground truth segment $q' \in \mathcal P_x$ as
\begin{equation}
    \IoU' (q',h) = \frac{| q' \cap Q' |}{| q' \cup Q' |}, \quad Q' = \bigcup_{\substack{ q \in \hat{\mathcal P}_x, q \cap q' \neq \emptyset \\ m(q) \leq h}} q \, .
\end{equation}
Thus, we obtain precision 
\begin{equation}
    \mathit{prec}(h) = \frac{\mathrm{TP}(h)}{\mathrm{TP}(h) + \mathrm{FP}(h)}
\end{equation}
and recall 
\begin{equation}
    \mathit{rec}(h) = \frac{\mathrm{TP}(h)}{\mathrm{TP}(h) + \mathrm{FN}(h)} 
\end{equation}
values on segment-level dependent of $h$.
The degree of separability is then computed as the area under precision recall curve ($\auc$) by thresholding the meta classification probability. 
Furthermore, we use the recall rate at $80\%$ precision rate ($\rec$) for the evaluation.
Moreover, we consider the segment-wise $F_{1}$ score which is defined by
\begin{equation}\label{eq:f1}
    F_{1} (h) = 2 \cdot \frac{\mathit{prec}(h) \cdot \mathit{rec}(h)}{\mathit{prec}(h) + \mathit{rec}(h)} \, .
\end{equation}
To obtain an evaluation metric independent of the meta classification threshold $h$, we calculate the averaged $F_{1}$ score 
\begin{equation}
    \bar{F_1} = \frac{1}{|H|} \sum_{h \in H} F_{1}(h)
\end{equation}
and the optimal $F_{1}$ score 
\begin{equation}
    F_1^* = \max_{h \in H} F_{1} (h) \, .
\end{equation}
%
%
\section{Effects of Meta Classification}\label{app:effect_meta_classif}
In \autoref{tab:meta_improve}, we show the effects of meta classification comparing the $F_{1}$ score (see \autoref{eq:f1}) performance with and without meta classification.  
$F_{1}(1)$ corresponds to the obtained precision and recall values without post-processing, i.e., meta classification and $F_1^*$ to the best possible ratio of both rates. Note, we use the meta classifier trained only on the source domain dataset Cityscapes. We observe that false positive pruning significantly improves the performance of our method as many false positive segments are predicted by the aggregation step to reduce the number of false negatives. We increase the $F_1$ score of up to $65.72$ pp for our method using meta classification. Noteworthy, the $F_1$ score for the basic semantic segmentation performance is also enhanced by up to $39.59$ pp. Moreover, the results show that without using meta classification the basic semantic segmentation prediction outperforms our method. This is caused by our foreground-background segmentation based on depth estimation being more prone to predicting foreground segments. We produce more possible foreground segments to reduce false negatives and using the false positive pruning, we outperform basic semantic segmentation.
%
%
\section{Numerical Results per Class}\label{app:results_class}
Up to now, the given results have been aggregated for all foreground classes, here we present results for three foreground classes separately, i.e., person, car and bicycle, see \autoref{tab:auprc_class}.
As the LostAndFound dataset provides only labels for road and small obstacles, a class-wise evaluation is not possible. In most cases, we outperform the basic semantic segmentation prediction, although differences for the datasets and the three classes are observed. The highest performance up to $89.20\%$ $\auc$ is achieved for Cityscapes since this is the source domain and thus, the semantic segmentation network produces strong predictions. Under domain shift, we obtain $\auc$ values of up to $75.53\%$. As for the foreground classes in general, there is no clear tendency which depth estimation network used in our method performs better.
For the class car, we achieve higher performance metrics in comparison to classes person and bicycle. Cars occur more frequently than persons and bicycles in all three datasets (see \cite{Cordts2016,Geyer2020,Varma2019}) and are easier to recognize given their larger size and similar shape.
In summary, we improve the detection performance of the basic semantic segmentation network in most cases and in particular, bridge the domain gap. Even though our performance for bicycles, for example, is comparatively lower, we generally detect more overlooked foreground segments and thus, reduce false negatives. 

\end{document}